\documentclass[10pt,twocolumn,letterpaper]{article}

\usepackage[pagenumbers]{iccv}
\usepackage[accsupp]{axessibility}

\definecolor{iccvblue}{rgb}{0.21,0.49,0.74}
\usepackage[pagebackref,breaklinks,colorlinks,allcolors=iccvblue]{hyperref}
\usepackage{pifont}
\usepackage{multirow}
\usepackage{xcolor}

\usepackage{changes}
\usepackage{algorithm}
\usepackage{algorithmicx}
\usepackage{algpseudocode}
\usepackage{amsmath}
\usepackage{tikz}

\usepackage{comment}
\usepackage{stfloats}
\def\paperID{9982} 
\def\confName{ICCV}
\def\confYear{2025}

\title{C$^2$MIL: Synchronizing Semantic and Topological Causalities in Multiple Instance Learning for Robust and Interpretable Survival Analysis}

\author{
Min Cen$^{1}$\thanks{Equal contribution. Min Cen contributed to this work during a visit at Xiamen University.}, Zhenfeng Zhuang$^{2}$\footnotemark[1], Yuzhe Zhang$^1$, Min Zeng$^1$, \\
Baptiste Magnier$^3$, Lequan Yu$^4$, Hong Zhang$^1$\thanks{Corresponding author.}, and Liansheng Wang$^2$\footnotemark[2]\\
$^1$University of Science and Technology of China, Hefei, China\\
$^2$Xiamen University, Xiamen, China\\
$^3$EuroMov Digital Health in Motion, Univ Montpellier, IMT Mines Ales, Ales, France\\
$^4$The University of Hong Kong, Pok Fu Lam, Hong Kong SAR, China\\
{\tt\small \{cenmin0127, zyz2020, zengm\}@mail.ustc.edu.cn, zhuangzhenfeng@stu.xmu.edu.cn,} \\
{\tt\small baptiste.magnier@mines-ales.fr, lqyu@hku.hk, zhangh@ustc.edu.cn, lswang@xmu.edu.cn}
}

\begin{document}
\maketitle
\begin{abstract}
Graph-based Multiple Instance Learning (MIL) is widely used in survival analysis with Hematoxylin and Eosin (H\&E)-stained whole slide images (WSIs) due to its ability to capture topological information. 
However, variations in staining and scanning can introduce semantic bias, while topological subgraphs that are not relevant to the causal relationships can create noise, resulting in biased slide-level representations. These issues can hinder both the interpretability and generalization of the analysis.
To tackle this, we introduce a dual structural causal model as the theoretical foundation and propose a novel and interpretable dual causal graph-based MIL model, C$^2$MIL. C$^2$MIL incorporates a novel cross-scale adaptive feature disentangling module for semantic causal intervention and a new Bernoulli differentiable causal subgraph sampling method for topological causal discovery. 
A joint optimization strategy combining disentangling supervision and contrastive learning enables simultaneous refinement of both semantic and topological causalities.  
Experiments demonstrate that C$^2$MIL consistently improves generalization and interpretability over existing methods and can serve as a causal enhancement for diverse MIL baselines. The code is available at \url{https://github.com/mimic0127/C2MIL}.
\end{abstract}
\section{Introduction}
\label{sec:intro}




\begin{figure}[t]
    \centering
    \includegraphics[scale=0.6]{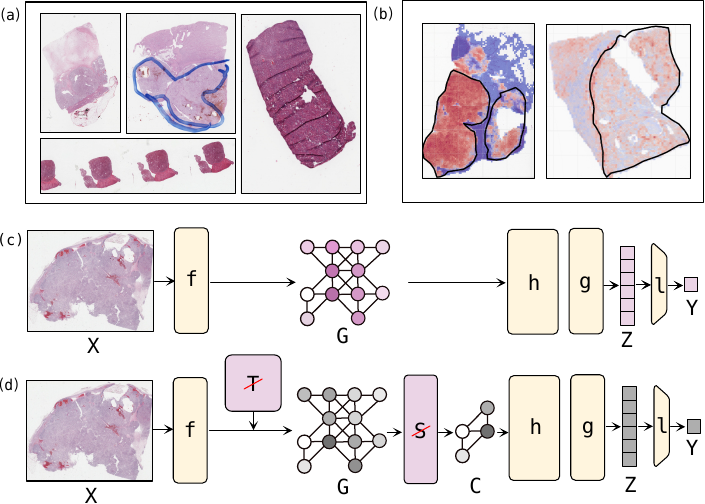}
    \caption{(a) Four WSIs showing staining, sectioning, and scanning variations. (b) Attention heatmaps from existing methods, revealing sensitivity to irrelevant regions (black outlines: tumor ground truth). (c) Standard graph-based MIL pipeline. (d) C$^2$MIL pipeline with semantic causal intervention for confounder adjustment and topological causal discovery for non-causal structure removal. Letters in (c) and (d) correspond to Section \ref{sec:causalper}.}
    \label{fig:intro}
\end{figure}

Hematoxylin and Eosin (H\&E)-stained whole slide images (WSIs) are the gold standard for pathological diagnosis, providing rich histological details essential for diagnostic and prognostic assessment. WSI-based survival analysis plays a vital role in predicting patient outcomes, guiding personalized treatment, and improving clinical decision-making \cite{zhu2017wsisa,liu2022deep}.  However, the sheer size and complexity of WSIs pose significant challenges for computational modeling. Multiple Instance Learning (MIL) has emerged as a powerful weakly supervised paradigm, reducing the need for labor-intensive manual slide review \cite{yao2020whole,li2021dual}. MIL has demonstrated outstanding performance in various tasks in pathology, including subtype classification \cite{hezi2025cimil}, gene mutation prediction \cite{ren2023iib}, and survival analysis \cite{liu2024advmil}.  Recently, graph-based MIL methods (Figure \ref{fig:intro} (c)) have gained traction for their ability to capture pathological features and model tissue and cellular topologies, further enhancing the predictive power of WSI-based survival analysis \cite{chen2021whole,cen2024orcgt,pal2022bag}.

Despite the success of MIL in survival analysis, trivial semantic feature bias in multi-institutional datasets remains a significant challenge \cite{song2023artificial}. 
As shown in Figure \ref{fig:intro}(a), variations frequently occur in slide preparation, such as staining protocols, tissue sectioning, and WSI scanning resolutions. 
Without bias correction, deep learning models may exploit irrelevant features, harming generalization. 
For instance, a model might rely on staining intensity rather than true histological characteristics to predict patient prognosis. 

Several recent methods have been proposed to address the challenge of semantic label-irrelevant features in histopathological images. Stain normalization \cite{macenko2009method,vahadane2016structure} standardizes color variations to improve generalizability but fails to account for biases from sectioning techniques and scanning resolutions. Contrastive learning-based augmentation \cite{ashraf2024enhancing,ke2021contrastive} enhances representation robustness but struggles to model variations from diverse slide preparation processes comprehensively. 
Beyond data-level preprocessing, causal inference-based MIL \cite{lin2024boosting,lin2023interventional,cuimultiscale} has been explored to mitigate semantic confounders. However, existing methods rely on multistage pipelines, increasing reproduction complexity and inefficiency. 
In addition, existing methods cluster patches independently, overlooking the shared trivial semantic features within a WSI, making patch-level clustering suboptimal. And these methods require a predefined cluster number ($K$), which varies by task and dataset, making it hard to choose the optimal $K$ without prior knowledge.

In addition to the difficulties encountered in managing trivial semantic features, the intricacies introduced by topological-level complexities further exacerbate the analytical process. Due to the high resolution of WSIs, only a small portion of their topological structure is causally relevant to clinical outcomes \cite{ibeling2021topological}. In graph-based MIL frameworks, irrelevant subgraphs introduce noise during patch-level aggregation, leading to biased slide-level representations. Therefore, identifying causal subgraphs is crucial for improving model interpretability and generalization. The majority of methodologies for the analysis of pathology images are predicated on graph attention mechanism; however, the integration of causal inference remains an underexplored area. Furthermore, the prevalent causality-driven graph models are predominantly tailored for classification, which renders them less appropriate for survival analysis.
To address these challenges, we introduce C$^2$MIL, a novel and interpretable dual-causal graph-based MIL framework that jointly models semantic and topological causalities through a dual structural causal model (SCM) (see Figure~\ref{fig:intro}~(d)).
C$^2$MIL performs semantic causal intervention by estimating semantic bias via cross-scale adaptive disentangling, enabling backdoor adjustment to remove trivial semantic confounders in patch representations.
This adaptivity has two core aspects: {\it i)} adaptively learning the cluster number without prior knowledge of institutional variations, and {\it ii)} adaptively identifying semantic confounders beyond staining bias.
For topological causal discovery, we propose a Bernoulli differentiable causal subgraph sampling method with a straight-through estimator (STE), ensuring efficient and robust structure learning.
Finally, we design a joint optimization strategy based on causal invariance, which combines semantic disentangling supervision and topological contrastive learning for simultaneous causal learning.
\begin{figure*}[htbp]
    \centering
    \includegraphics[width=0.9\textwidth]{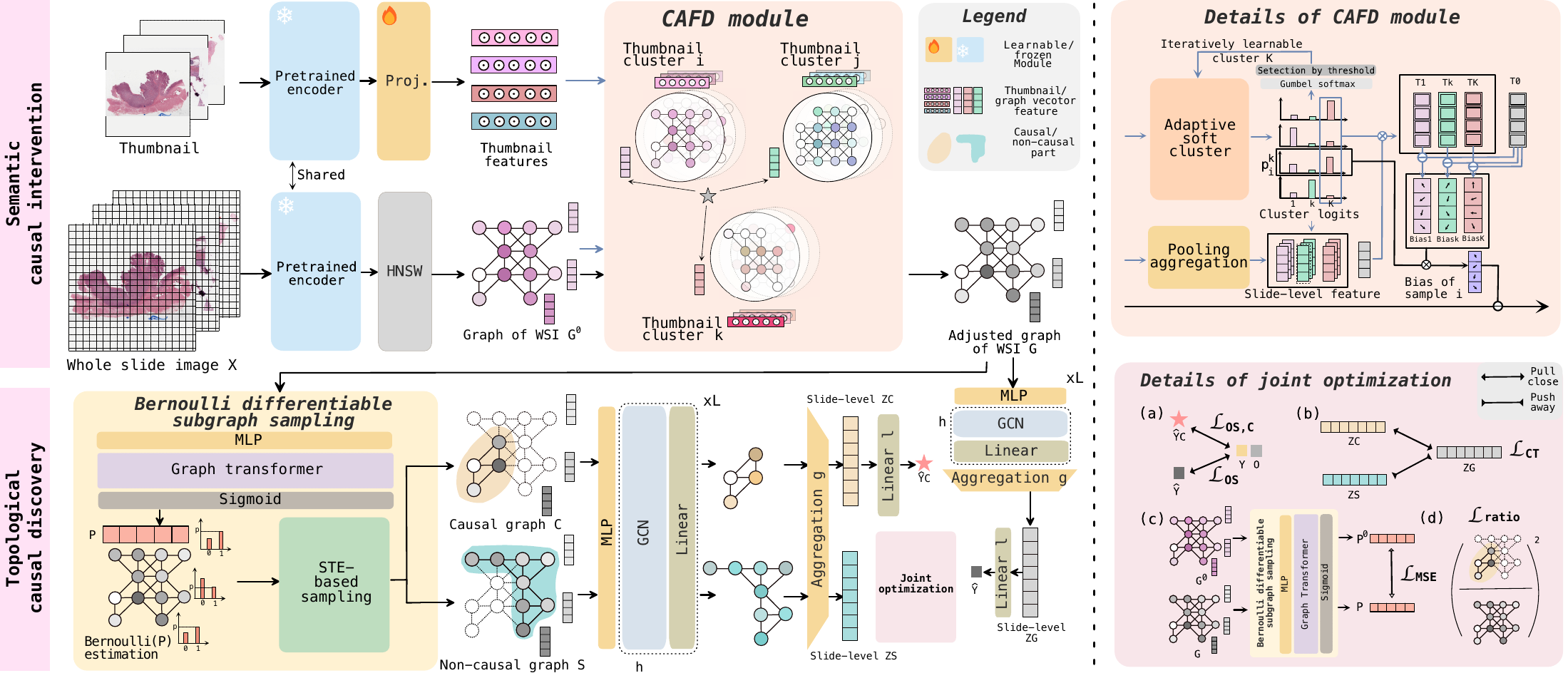}
    \caption{C$^2$MIL comprises semantic causal intervention and topological causal discovery. 
    Cross-scale adaptive disentangling integrates thumbnails and patches for semantic intervention, with Bernoulli differentiable subgraph sampling enabling topological discovery.
    A joint optimization strategy is proposed to enable end-to-end training.}
    \label{fig:overview}
\end{figure*}

Our main contributions are as follows:
\begin{itemize}
\item We propose C$^2$MIL, a theoretically grounded dual-causal graph-based MIL model that removes semantic confounders and discovers causal sub-topologies, enhancing accuracy, generalizability, and interpretability.
\item We introduce cross-scale adaptive feature disentangling (CAFD) for semantic intervention via backdoor adjustment, autonomously estimating confounders without prior knowledge.
\item We develop a Bernoulli differentiable causal subgraph sampler with STE adjustment, enabling robust causal topology learning within a graph transformer.
\item We design a joint optimization strategy that unifies semantic disentangling and topological contrastive learning under a causal invariance principle.
\item Extensive experiments on three public datasets demonstrate state-of-the-art performance, with interpretable clustering and attention visualizations.
\end{itemize}

\section{Related Work}
\label{sec:relatedwork}

\subsection{Multiple Instance Learning}

Multiple Instance Learning (MIL) is the dominant approach for gigapixel-level WSI analysis, treating each WSI as a bag of patches with bag-level labels. Due to the huge size of WSIs, end-to-end training is impractical, so MIL typically follows the two-stage pipeline: (1) feature extraction using pretrained models such as ViT \cite{dosovitskiy2020image}, ctranspath \cite{wang2022transformer}, or UNI \cite{chen2024towards} and (2) feature aggregation and prediction. For aggregation, ABMIL \cite{ilse2018attention} uses self-attention for interpretable patch weighting, while DTFDMIL \cite{zhang2022dtfd} mitigates sample scarcity via pseudo-bags and feature distillation. RRTMIL \cite{tang2024feature} enhances MIL by re-embedding instance features to capture fine-grained local information. Recently, graph-based MIL has gained traction by modeling cell and tissue topologies. PatchGCN \cite{chen2021whole} captures spatial relationships for better classification. Li \etal. \cite{Li_2024_CVPR} propose a dynamic graph for WSIs that uses knowledge graphs with adaptive neighbor embeddings and knowledge-aware attention to refine node features and improve classification. These highlight the growing impact of graph-based MIL in WSI analysis. 

\subsection{Causal Inference}

Causal inference enhances interpretability, identifies causal mechanisms, and mitigates confounders, improving robustness in deep learning. Sui \etal. \cite{sui2022causal} introduced Causal Attention Learning (CAL) to remove shortcut features, while Zhao \etal. \cite{zhao2024causality} developed a causality-driven generative model for DyGNNs. In pathological image analysis, IBMIL \cite{lin2023interventional} used backdoor adjustment to correct bag-level bias, and CaMIL \cite{chen2024camil} applied front-door adjustment for WSI classification. However, existing methods lack causal inference for graph-based MIL and rely on multi-step pipelines, increasing complexity. In contrast, our method simultaneously eliminates confounders and discovers causal structures in a unified learning process, making it more efficient and offering a comprehensive causal perspective at the slide level.
\section{Methodology}
\label{sec:methodology}
Figures \ref{fig:overview} and \ref{fig:scm} show the overall C$^2$MIL framework and the dual causal structural model (DSCM). C$^2$MIL enhances graph-based MIL by explicitly modeling both semantic and topological causalities. Specifically, it uses a cross-scale adaptive feature disentangling (CAFD) module for backdoor adjustment of semantic features and a differentiable causal subgraph sampler for robust topology discovery. A unified optimization strategy ensures that both semantic and topological components are learned jointly under the causal invariance principle.

\subsection{Causal Perspective of C$^2$MIL}
\label{sec:causalper}

\noindent \textbf{Graph-based MIL formulation.}
As shown in Figure~\ref{fig:intro}(c), a WSI is treated as a bag $X = \{x_1, x_2, \dots, x_n\}$ with survival outcome $(t, O)$, where $t$ is the observed time and $O$ the censoring indicator. A pretrained extractor $f$ generates patch features $V = {v_1, v_2, \dots, v_n}$ as nodes to form a graph $G = \{V, E\}$, where $E = \{e_{ij}\}$ encodes topological relations ($e_{ij} \in {0,1}$).
Graph-based MIL for survival analysis models the relationship between $G$ and the event risk. A typical framework includes a graph feature learner $h$, a bag-level aggregator $g$, and a risk predictor $l$, yielding $Z = g(h(G))$ and predicted hazard $\hat{Y} = l(Z)$. The objective is to learn a risk function under censoring, typically with the partial likelihood of the Cox model~\cite{katzman2018deepsurv}.

\begin{figure}[t]
    \centering
    \includegraphics[width=0.42\textwidth]{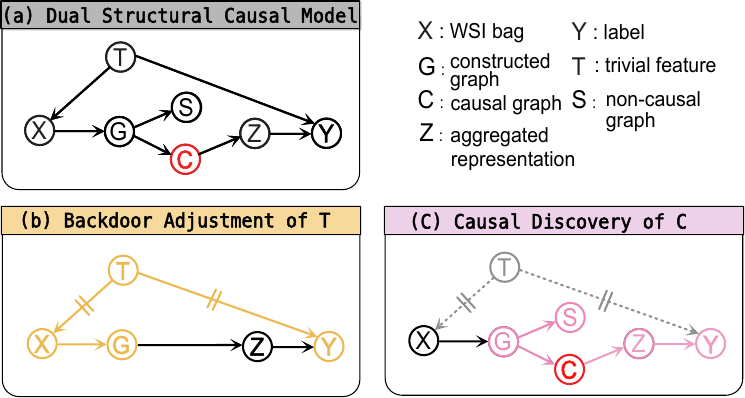}
    \caption{The structure of C$^2$MIL consisting of (a) structural causal model (SCM) and two dual causality modules, \textit{i.e.}, (b) backdoor adjustment of $T$ and (c) causal discovery of $C$.}
    \label{fig:scm}
\end{figure}

\noindent \textbf{Dual structural causal models (SCM) for C$^2$MIL.}
To synchronize semantic and topological causalities, we introduce a dual structural causal model $\mathcal{G}$ for C$^2$MIL, as shown in Figure \ref{fig:scm}. 
In addition to previously defined symbols, $T$ represents trivial semantic features ({\it e.g.}, staining differences and institutional variations), $C$ denotes the causal topological subgraph in $G$ that causally influences $Y$, and $S$ represents non-causal subgraphs within instances. 
There are four key causal relationships: (1) $X \leftarrow T \rightarrow Y$: Trivial features $T$ (e.g., WSI color, slide count per WSI), due to various preparation methods, influence both WSI $X$ and label $Y$; (2) $X \rightarrow G$: The graph $G$ is constructed from $X$ and thus causally influenced by $X$; (3) $S \leftarrow G \rightarrow C$: Causal and no-causal subgraphs $C$ (e.g., tumor) and $S$ (e.g., stroma or background)) are derived from $G$, therefore determined by G; (4) $C \rightarrow Z \rightarrow Y$: Features $Z$ aggregated from $C$ causally influence the label $Y$.

\noindent \textbf{Semantic causal intervention and topological causal discovery.} In the SCM, a backdoor path $G \leftarrow X \leftarrow T \rightarrow Y$ exists, introducing semantic confounder $T$. To estimate $P(Y \mid \text{do}(G))$, we apply backdoor adjustment, blocking the semantic confounding effect of $T$:
\begin{equation}
\begin{split}
P(Y \mid \text{do}(G)) 
= \sum_t P(Y \mid \text{do}(G), T=t) \, P(T=t \mid \text{do}(G)) \\
= \sum_t P(Y \mid \text{do}(G), T=t) \, P(T=t)
\end{split}
\label{eq:backdoor1}
\end{equation}

In addition, $G$ contains a causal subgraph $C$ and a non-causal subgraph $S$. We further estimate $P(Y \mid \text{do}(G), t)$ by performing causal discovery on topological subgraph $C$, and therefore enhance the generalizability and interpretability of this model:

\begin{equation}
\small
\begin{split}
P(Y \mid \text{do}(G),t) &= \sum_c P(Y\mid \text{do}(G),c,t)P(c\mid\text{do}(G),t) \\
&=\sum_c P(Y\mid c, t)P(c\mid\text{do}(G))\\
&=\sum_c P(Y\mid \text{do}(C=c), t)P(c\mid G).
\label{eq:backdoor2}
\end{split}
\end{equation}
Combining equations \eqref{eq:backdoor1}-\eqref{eq:backdoor2}, we obtain:
\begin{equation}
\resizebox{0.9\hsize}{!}{$
\begin{split}
P(Y\mid \text{do}(G)) = \sum_t P(t) \sum_c P(Y\mid \text{do}(C=c),t)P(c\mid G).
\end{split}
$}
\label{eq:eqall}
\end{equation}
Thus, we first estimate $P(T)$ and apply backdoor adjustment to obtain causal semantic features. Then, we estimate $P(C\mid G)$ via causal discovery, identifying the causal subgraph $C$. To ensure causal invariance, fidelity, and interpretability, $C$ must retain label-relevant information while preserving consistency with the original input and we should remove the influence of semantic confounder $T$ and non-causal topological subgraph $S$.

\subsection{Cross-scale Adaptive Feature Disentangling}



As formulated in equation \eqref{eq:eqall}, we estimate $P(T)$ to account for semantic confounders. To this end, we propose a Cross-scale Adaptive Feature Disentangling (CAFD) module, which removes trivial semantic features $T$ while extracting causally relevant ones through backdoor adjustment. Given that patches from the same slide share similar trivial features due to uniform preparation, we leverage the multi-scale nature of pathology images — where smaller scales capture global semantic information like stain color and larger scales provide fine-grained details like tissue morphology — to adaptively disentangle features and estimate $P(T)$ without prior slide processing knowledge. 

Let the slide-level thumbnail $i$ be $\tau_i$. A pretrained feature extractor $F$ extracts the thumbnail features $f_i$ and patch features $V_i=\{v_{i1}, v_{i2}, \dots, v_{im}\}$ for $m$ patches in a slide. 
A projection (Proj) layer learns trivial features $t_i$ from thumbnail, capturing staining and preparation variations: $t_i = \text{Proj}(f_i) = \text{MLP}(F(\tau_i))$.
To estimate $P(T)$, we apply soft $K$-means clustering to $\{t_i\}$, assigning each sample to cluster groups $\{Gr_1, Gr_2, \dots, Gr_K\}$ while preserving gradient backpropagation. The probability of $t_i$ belonging to cluster $Gr_k$, denoted as $p_{i}^{k}$, is computed as:
\begin{equation}
\small
p_i^k = P(t_i\in Gr_k\mid \tau _i) = \text{SoftKmeans}_K(\text{MLP}(F(\tau_i)), 
\end{equation}
where $k \in \{1,\dots,K\}$. Selecting the optimal number of cluster centers is challenging. To address this, we propose a novel dynamic iterative approach using Gumbel-Softmax \cite{jang2016categorical}, which adaptively approximates the effective number of clusters $K_{\text{effective}}$. 
Specifically, the soft weight $w^k$ for each cluster $k$ is computed using Gumbel-Softmax on learnable cluster logits $s^k$, which then allows us to estimate the effective number of clusters $K_{\text{effective}}$: 
\begin{align}
\small
 &w^k = \text{GumbelSoftmax}_i \left(\frac{\log(s^k) + \epsilon_k}{\tau}\right),\\
&K_{\text{effective}} = \sum_{k=1}^{K_{\max}} \mathbb{I}\left( w^k > 0.1 \right),
\end{align}
where $\epsilon_k$ is the Gumbel noise, $\tau$ is a temperature parameter, and $K_{\text{max}}$ is a predefined maximum number of clusters.
$K_{\text{effective}}$ is used for K-means++ \cite{arthur2006k} initialization and is dynamically updated at each iteration of the whole pipeline, rather than within the soft $K$-means process.

Based on the clustering logits prediction of the slide-level, the estimation of $P(T)$ under the patch feature distribution can be obtained. The semantic confounding feature of the  $k$-th cluster $Gr_k$ is computed as:
\begin{equation}
T_k = \frac{1}{\sum_i p_i^k} \sum_i p_i^k \frac{1}{N_i} \sum_j v_{ij},
\end{equation}
where $N_i$ is the number of patches in sample $i$. The global distribution of trivial features for the across the dataset is:
\begin{equation}
T_0 = \frac{1}{N} \sum_i \frac{1}{N_i} \sum_j v_{ij},
\end{equation}
where $N$ is the sample size.
To mitigate computational challenges from excessive patches, we randomly sample $n$ instances per sample to estimate the distribution of semantic confounding features $T$. The bias for the $k$-th distribution is then:
\begin{equation}
 \text{Bias}_k = T_k - T_0.
\end{equation}
For each sample $i$, we remove this bias from each patch feature to obtain semantic confounder-free features $\tilde{v}_{ij}$:
\begin{equation}
\tilde{v}_{ij} = v_{ij} - \sum_{k} p_i^k \, \text{Bias}_k.
\end{equation}
Thus, we obtain the graph $G_i = (\tilde{V}_i, E_i)$ with causal semantic information, which is free from the semantic confounder $T$, where $\tilde{V}_i=\{\tilde{v}_{i1}, \tilde{v}_{i2}, \dots, \tilde{v}_{im}\}$.

\subsection{Bernoulli Differentiable Subgraph Sampling}
Following equation \eqref{eq:eqall}, after semantic causal intervention, we estimate $P(C \mid G)$ to identify the topological causal subgraph $C$. 
This transforms $P(Y \mid \text{do}(G))$ into $P(Y \mid \text{do}(C), T)$, enhancing interpretability and generalization. Let $C_i$ be the $i$-th subgraph, and $o_{ij}=1$ if the $j$-th patch is included in $C_i$ and $o_{ij}=0$ otherwise. The likelihood of $C_i$ is
\begin{equation}
\label{eq:PCG}
\small
P(C_i | G_i) = \prod_{j=1}^{|G_{i}|} p_{ij}^{o_{ij}} (1 - p_{ij})^{1 - o_{ij}},
\end{equation}
where $o_{ij}\sim \text{Bernoulli}(p_{ij})$, $i.e.$, $p_{ij}=P(o_{ij}=1|G_i)$.
To capture global topological causal dependencies, an MLP layer and a two-layer Graph Transformer (GT) \cite{yun2019graph} are used to estimate $P(C_i|G_i)$, where the MLP first projects $G$ to a lower-dimensional space $G' := \text{MLP}(G)$ to reduce computation. The edge attribute $A_i$$=$$\{a_{jk}\}_{{j,k}}$$=$$\{[\tilde{v}_{ij},\tilde{v}_{ik}]\}_{j,k\in[1,N_i]}$ in $G_i$ is obtained by concatenating the features of the two endpoint nodes. Then,
\begin{equation}
    p_{ij} = \sigma(\text{GT}(\text{MLP}(\tilde{V}_i),E_i,A_i)[j]),
\end{equation}
where $[j]$ denotes the $j$-th index of the output node logits $\text{GT}(\text{MLP}(\tilde{V}_i),E_i,A_i)$, $\sigma$ is the sigmoid function.

Next, we randomly sample all causal subgraphs $C_i$ using the estimated Bernoulli distribution $p_{ij}$, thereby obtaining the causal complementary subgraphs $C_i$ and $S_i$, {\it i.e.}, $G_i = C_i \cup S_i$ and $C_i \cap S_i = \phi$. The advantage of using random sampling instead of directly multiplying the predicted soft mask $P=\{p_{ij}\}_{j \in [1, N_i]}$ by $G_i$ is that it prevents excessive smoothing and allows nondeterministic causal subgraph selection, improving generalization and reducing overfitting.

During training, to enable gradient backpropagation, we apply the straight-through estimator \cite{bengio2013estimating} adjustment, which enables hard mask sampling and selection of causal subgraph $C_i$. For model validation and testing, we apply $P$ as a soft mask on the graph $G$. According to the law of large numbers, the sample mean will almost surely converges to the true probability $P$. The pseudocodes of this module are provided in Section \ref{sup:ssp} of Supplementary Materials.

\subsection{Network Design and Joint Optimization}

\begin{table*}[htbp]
  \centering
  \small
  \renewcommand{\arraystretch}{1.2}
  \setlength{\tabcolsep}{3pt}
  \resizebox{0.95\textwidth}{!}{
  \begin{tabular}{l|cc|ccc|ccc}
    \toprule
    \multirow{3}{*}{Model} & \multicolumn{2}{c|}{Strategies} & \multicolumn{3}{c|}{Five-fold cross validation} & \multicolumn{3}{c}{Out-of-distribution external validation} \\
    \cmidrule(lr){2-3} \cmidrule(lr){4-6} \cmidrule(lr){7-9}
    & Graph & Causal & TCGA-KIRC & TCGA-ESCA & TCGA-BLCA & TCGA-KIRC & TCGA-ESCA & TCGA-BLCA \\
    \midrule
    ABMIL \cite{ilse2018attention} & \ding{55} & \ding{55} & 0.6794$_{0.0441}$ & 0.6385$_{0.0622}$ & 0.5771$_{0.0229}$ & 0.5971$_{0.0419}$ & 0.6143$_{0.0182}$ & 0.6728$_{0.0472}$ \\
    TransMIL \cite{shao2021transmil} & \ding{55} & \ding{55} & 0.6658$_{0.0602}$ & 0.5651$_{0.0305}$ & 0.5680$_{0.0427}$ & 0.6103$_{0.0235}$ & 0.5393$_{0.0351}$ & 0.6762$_{0.0435}$ \\
    RRTMIL \cite{tang2024feature} & \ding{55} & \ding{55} & 0.6775$_{0.0594}$ & 0.6196$_{0.0412}$ & 0.5661$_{0.0331}$ & 0.5842$_{0.0465}$ & 0.5893$_{0.0639}$ & 0.6786$_{0.0424}$ \\
    DeepGraphConv \cite{li2018graph} & \ding{52} & \ding{55} & 0.6674$_{0.0245}$ & 0.6118$_{0.0908}$ & 0.5720$_{0.0229}$ & 0.5091$_{0.0473}$ & 0.5982$_{0.0587}$ & 0.6130$_{0.0772}$ \\
    PatchGCN \cite{chen2021whole} & \ding{52} & \ding{55} & 0.6858$_{0.0261}$ & 0.6519$_{0.0562}$ & 0.5757$_{0.0201}$ & 0.6057$_{0.0289}$ & 0.5679$_{0.0421}$ & 0.6974$_{0.0111}$ \\
    ProtoSurv \cite{wu2025leveraging} & \ding{52} & \ding{55} & 0.6975$_{0.0490}$ & 0.6194$_{0.0540}$ & 0.5926$_{0.0320}$ & 0.6098$_{0.0407}$ & 0.5982$_{0.0421}$ & 0.6951$_{0.0485}$ \\
    IBMIL \cite{lin2023interventional} & \ding{55} & \ding{52} & 0.6970$_{0.0541}$ & 0.5893$_{0.0468}$ & 0.5527$_{0.0320}$ & 0.6163$_{0.0232}$ & 0.5714$_{0.0677}$ & 0.6537$_{0.0398}$ \\
    \midrule
    C$^2$MIL (Ours) & \ding{52} & \ding{52} & \textbf{0.7078}$_{0.0512}$ & \textbf{0.6904}$_{0.0744}$ & \textbf{0.6081}$_{0.0417}$ & \textbf{0.6275}$_{0.0225}$ & \textbf{0.6500}$_{0.0428}$ & \textbf{0.7015}$_{0.0386}$ \\
    \bottomrule
  \end{tabular}
  }
  \caption{Performance comparison of different models in terms of C-index averaged over five-fold cross validation or out-of-distribution external validation. The subscripts represent standard deviations.}
  \label{tab:combined_results}
\end{table*}

\begin{figure*}[htbp]
    \centering
    \includegraphics[width=\textwidth]{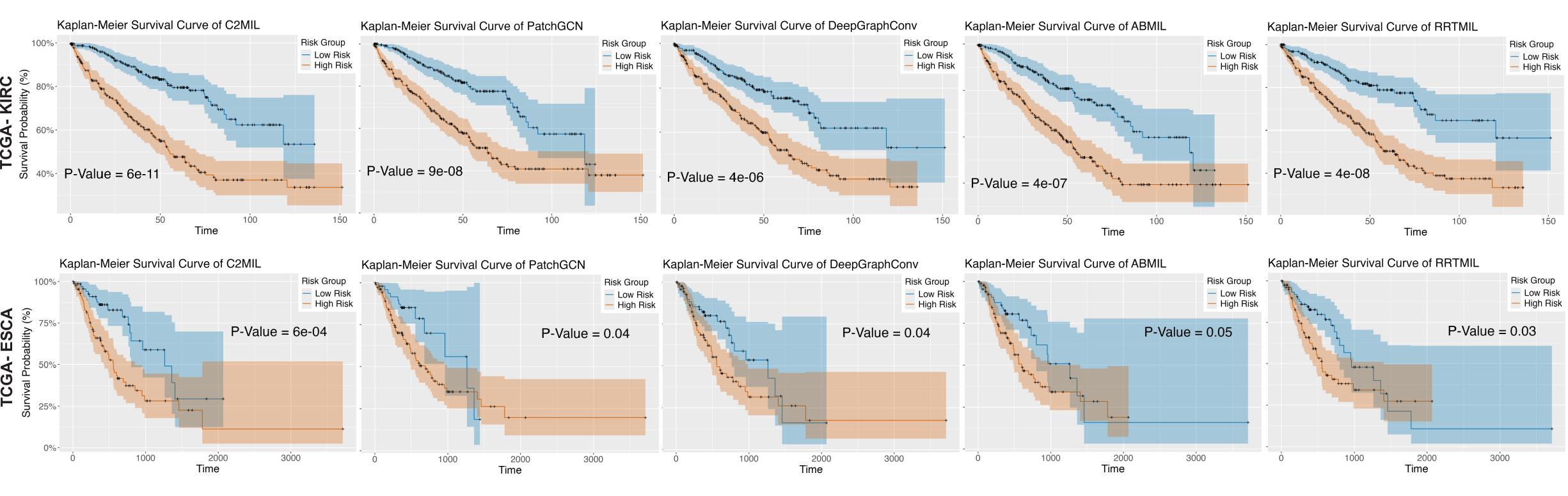}
    \caption{Kaplan-Meier curves of low risk and high risk patients by four methods (C$^2$MIL, PatchGCN, DeepGraphConv, ABMIL, RRTMIL). P-values of log-rank tests for comparing two curves are also presented.}
    \label{fig:KM}
\end{figure*}

Based on causal invariance, as well as fidelity and interpretability, we design our network and propose a joint optimization strategy to achieve the simulation of semantic trivial feature intervention and topological causal discovery.
Specifically, we define the graph constructed using the original patches with features $V$ for the $i$-th sample as $G^0_i$. Then $G_i$$ =$$ \text{SemanticIntervention}(G^0_i, \tau_i)$. 
We obtain causal subgraph $C_i$ and non-causal subgraph $S_i$ by $\text{CausalTopologicalSample}(G_i)$. In our network, $G_i$, $C_i$, and $S_i$ are then passed through $h$ and $g$, producing $Z_{Gi} = h(G_i)$, $Z_{Ci} $$= $$h(C_i)$, and $Z_{Si} $$=$$ h(S_i)$. Finally, the label prediction results are obtained: $\hat{Y}_{i} $$=$$ l(Z_{Gi}), \hat{Y}_{Ci} $$=$$ l(Z_{Ci})$. During inference and evaluation, the final prediction is $\hat{Y}_{Ci}$.

To simultaneously optimize the estimation and refinement of semantic and topological causalities, we propose a joint optimization objective that includes four components: optimization of survival analysis performance, trivial semantic feature disentangling, causal graph contrastive mechanism, and causal ratio loss.

\noindent \textbf{Survival analysis optimization.} To optimize the performance of predictions of $\hat{Y}_C$ and $\hat{Y}$ in survival analysis, we use Cox loss functions (CoxLoss) \cite{ranganath2016deep} to improve model's survival time prediction. The overall survival (OS) labels for sample $i$ consist of the survival time $t_i$ and the censoring status $O_i$. First, define Cox loss functions as follows:
\begin{equation}
\resizebox{0.87\hsize}{!}{$
\left\{
\begin{split}
& \mathcal{L}_{{\text{OS}}, \text{C}} = - \sum_{i: O_i = 1} \bigg[ \hat{Y}_{Ci} - \log \sum_{j \in R(t_i)} \exp(\hat{Y}_{Cj})  \bigg], \\
& \mathcal{L}_{\text{OS}} = - \sum_{i: O_i = 1} \bigg[ \hat{Y}_{i} - \log \sum_{j \in R(t_i)} \exp(\hat{Y}_{j})  \bigg],
\end{split}
\right.
$}
\end{equation}
where $\hat{Y}_{Ci}$ represents the predicted risk for sample $i$, and $R(t_i)$ is the set of samples with survival times greater than or equal to $t_i$.

\noindent \textbf{Semantic trivial feature disentangling.} Based on causal invariance, the semantic information in the graph that is causally related to the label should remain unchanged after adjusting for trivial semantic features. Therefore, we expect the estimated models $P(C_i | G_i)$ and $P(C_i | G_i^0)$ to be as close as possible. According to equation \eqref{eq:PCG}, we optimize this with mean square error (MSE) loss:
\begin{equation}
\resizebox{0.87\hsize}{!}{$
\begin{split}
\mathcal{L}_{\text{MSE}} = \frac{1}{N} \sum_{i=1}^{N} \frac{1}{N_i} \sum_{j=1}^{N_i} ( P(o_{ij} = 1 | G_i) - P(o_{ij} = 1 | G_i^0) )^2 ,
\end{split}
$}
\end{equation}
where $N$ is the total number of samples and $N_i$ is the number of instances in the $i$-th sample.

\noindent \textbf{Causal topological graph contrasive mechanism.} To ensure causal fidelity and interpretability, the sampled causal subgraph features $Z_{C_i}$ should align with the original input features $Z_{G_i}$, while irrelevant subgraph features $Z_{S_i}$, treated as noise, should remain distant. To enforce this, we introduce a contrastive (CT) mechanism based on slide-level features, with the following contrastive loss:
\begin{equation}
\small
\mathcal{L}_{\text{CT}} = -\frac{1}{N} \sum_{i=1}^{N}\log \left( \frac{\exp( u_i / \nu)}{\exp( u_i / \nu) + \exp( v_i / \nu)} \right),
\end{equation}
where $u_i=\text{cos}(Z_{Gi}, Z_{Ci})$, $v_i=\text{cos}(Z_{Gi}, Z_{Si})$,
$N$ is the number of samples, and $\nu$ is the temperature parameter used to control the smoothness of the softmax function.

\noindent \textbf{Causal ratio control.} To reduce label-irrelevant subgraph influence, the causal subgraph should be minimal. To enforce this, we introduce a causal subgraph ratio loss, penalizing larger sampled subgraphs. Thus, we define
\begin{equation}
\small
\mathcal{L}_{\text{Ratio}} = \left(\frac{1}{N} \sum_{i=1}^{N} \frac{|C_i|}{|G_i|} \right)^2,
\end{equation}
where $|C_i|$ represents the number of nodes in $C_i$.

For C$^2$MIL, the final joint optimization objective is
\begin{equation}
\small
\mathcal{L} = \mathcal{L}_{{\text{OS}}, \text{C}}+\mathcal{L}_{\text{OS}}+\lambda_1\mathcal{L}_{\text{MSE}}+\lambda_2\mathcal{L}_{\text{CT}}+\lambda_3\mathcal{L}_{\text{Ratio}},
\end{equation}
where $\lambda_1$, $\lambda_2$, $\lambda_3$ represent  three hyperparameters. 

\section{Experiments}
\label{sec:experiments}

\subsection{Experimental Setup}

\noindent \textbf{Datasets.} Three publicly available datasets from The Cancer Genome Atlas (TCGA) \cite{weinstein2013cancer}, including \textbf{TCGA-KIRC} ($N=512$), \textbf{TCGA-ESCA} ($N=155$), and \textbf{TCGA-BLCA} ($N=385$), are used to evaluate the survival prediction performance of our method. These datasets consist of samples collected from multiple institutions. To assess the generalization ability of our model, we randomly selected data from a single institution in each dataset as an external validation set, resulting in three independent validation subsets: \textbf{TCGA-KIRC-CJ} ($N=70$), \textbf{TCGA-ESCA-L5} ($N=20$), and \textbf{TCGA-BLCA-FD} ($N=48$).

\noindent \textbf{Evaluation metrics.} We evaluated the prognostic performance of survival analysis using the concordance index (C-index) \cite{harrell1984regression}. Additionally, we plotted Kaplan-Meier \cite{kaplan1958nonparametric} curves and performed the log-rank test on Cox proportional hazard model \cite{lin1989robust} to obtain p-values for differences in high- and low-risk populations predicted by the model. To mitigate the impact of randomness of dataset partitioning, we conducted experiments using five-fold cross-validation and out-of-distribution experiments.

\noindent \textbf{Implementation details.} Implementation details are given in Section \ref{sup:Implement} of Supplementary Material.

\noindent \textbf{Comparison methods.} We compared our model C$^2$MIL with seven other advanced models: ABMIL \cite{ilse2018attention}, TransMIL \cite{shao2021transmil}, IBMIL \cite{lin2023interventional}, RRTMIL \cite{tang2024feature}, DeepGraphConv \cite{li2018graph}, PatchGCN \cite{chen2021whole}, and ProtoSurv \cite{wu2025leveraging}. IBMIL is a MIL based on set with causal intervention, and ProtoSurv is based on heterogeneous graph.
\begin{table}[t!]
  \centering
  \renewcommand{\arraystretch}{1.2}
  \setlength{\tabcolsep}{1.5pt}
  \resizebox{0.48\textwidth}{!}{
  \begin{tabular}{l|ccc}
    \toprule
    Model & TCGA-KIRC & TCGA-ESCA & TCGA-BLCA \\
    \midrule
    ABMIL & 0.6794 & 0.6385 & 0.5771 \\
    ABMIL + semantic & 0.6996$_{\textcolor{ForestGreen}{+0.0202}}$ & 0.6266$_{\textcolor[RGB]{150,0,0}{-0.0119}}$ & 0.6023$_{\textcolor{ForestGreen}{+0.0252}}$ \\
    \midrule
    TransMIL & 0.6658 & 0.5651 & 0.5680 \\
    TransMIL + semantic & 0.6818$_{\textcolor{ForestGreen}{+0.0160}}$ & 0.6368$_{\textcolor{ForestGreen}{+0.0717}}$ & 0.5773$_{\textcolor{ForestGreen}{+0.0093}}$ \\
    \midrule
    DeepGraphConv & 0.6674 & 0.6118 & 0.5720 \\
    DeepGraphConv + topology & 0.6814$_{\textcolor{ForestGreen}{+0.0140}}$ & 0.6562$_{\textcolor{ForestGreen}{+0.0444}}$ & 0.5667$_{\textcolor[RGB]{150,0,0}{-0.0053}}$ \\
    DeepGraphConv + semantic + topology & 0.6760$_{\textcolor{ForestGreen}{+0.0086}}$ & 0.6711$_{\textcolor{ForestGreen}{+0.0593}}$ & 0.5841$_{\textcolor{ForestGreen}{+0.0121}}$ \\
    \midrule
    ProtoSurv & 0.6975 & 0.6194 & 0.5926  \\
    ProtoSurv + topology & 0.6998$_{\textcolor{ForestGreen}{+0.0023}}$ & 0.6279$_{\textcolor{ForestGreen}{+0.0085}}$ & 0.5826$_{\textcolor[RGB]{150,0,0}{-0.0100}}$ \\
    ProtoSurv + semantic + topology & 0.7048$_{\textcolor{ForestGreen}{+0.0073}}$ & 0.6512$_{\textcolor{ForestGreen}{+0.0318}}$ & 0.6064$_{\textcolor{ForestGreen}{+0.0138}}$ \\
    \midrule
    PatchGCN & 0.6858 & 0.6519 & 0.5757 \\
    PatchGCN + topology & 0.6926$_{\textcolor{ForestGreen}{+0.0068}}$ & 0.6788$_{\textcolor{ForestGreen}{+0.0269}}$ & 0.5844$_{\textcolor{ForestGreen}{+0.0087}}$ \\
    PatchGCN + semantic + topology & 0.7078$_{\textcolor{ForestGreen}{+0.0220}}$ & 0.6904$_{\textcolor{ForestGreen}{+0.0385}}$ & 0.6081$_{\textcolor{ForestGreen}{+0.0324}}$ \\
    \bottomrule
  \end{tabular}
  }
  \caption{Analysis of causal modules with multiple baselines. The subscripts are the increments compared against the  baselines.}
  \label{tab:model_comparison}
\end{table}

\subsection{Predictive Performance Comparison}

\noindent \textbf{Internal Cross Validation.}
As reported in Table~\ref{tab:combined_results}, our method achieves state-of-the-art performance in five-fold cross-validation across three TCGA cohorts, with mean C-index of 0.7078, 0.6904, and 0.6081, respectively. C$^2$MIL outperforms existing methods by 2.20-4.04\%, 3.85-12.53\%, and 3.10-5.54\% on these datasets, including a 1.08-10.11\% improvement over the causal baseline IBMIL.

Figure \ref{fig:KM} illustrates the Kaplan-Meier survival curves and the log-rank test result of Cox proportional hazards for different models on the test set of TCGA-ESCA and TCGA-KIRC, and the predicted high and low risks are determined by the median risk values predicted in the training set. 
C$^2$MIL achieves the most distinct separation between the high-risk and low-risk groups ($p$-value$=$$6\times 10^{-4}$ in TCGA-ESCA; $p$-value$=$$6\times 10^{-11}$ in TCGA-KIRC).

\noindent \textbf{Out-of-distribution Generalization.} For out-of-distribution (OOD) validation, each primary dataset (TCGA-KIRC-CJ, TCGA-ESCA-L5, TCGA-BLCA-FD) is used as an independent external validation set. The remaining data underwent five-fold cross-validation. Table \ref{tab:combined_results} shows cross-validated models' average performance on these external validation sets.

Notably, C$^2$MIL exhibits strong generalizability, achieving C-indexes of 0.6275, 0.6500, and 0.7015 on external datasets, outperforming baselines from 2.86\% to 8.26\% on average. 

\subsection{Ablation Study and Hyperparameter Analysis}

\noindent \textbf{Analysis of C$^2$MIL Components.}
The proposed modules can be applied to any graph-based MIL. 
CAFD module is compatible with any MIL model. To evaluate the effectiveness and generalizability of these modules in C$^2$MIL, we integrated them into various models and conducted ablation experiments. For graph-based models, we ablated both the topological causal discovery network and the adaptive disentangling module, while for other models, we examined the latter. 
The variations in performance, indicated by the increments and decrements in the lower-right corner of the results, are compared against the respective baselines.

As shown in Table \ref{tab:model_comparison}, ablation experiments across multiple baselines demonstrate that the modules designed in C$^2$MIL consistently and significantly improve baseline performance. Specifically, the semantic disentangling module alone achieves a maximum improvement of 2.52\%, highlighting the importance of reducing domain bias, while the graph causal network module alone leads to a maximum improvement of 2.37\%, demonstrating the effectiveness of identifying causal substructures. When combined, these two modules yield an improvement of up to 5.93\%, underscoring their complementary effectiveness and generalizability in enhancing the model. 

\begin{table}[htbp]
  \centering
  \renewcommand{\arraystretch}{1.2}
  \setlength{\tabcolsep}{2pt}
  \resizebox{0.5\textwidth}{!}{
      \begin{tabular}{l|ccc}
        \toprule
        Model & TCGA-KIRC & TCGA-ESCA & TCGA-BLCA \\
        \midrule
        C$^2$MIL w/o Disentangling Loss & 0.7042 & 0.6542 & 0.6022 \\
        C$^2$MIL w/o Contrasive Loss & 0.6779 & 0.6252 & 0.5783 \\
        C$^2$MIL w/o Ratio Loss & 0.6914 & 0.6702 & 0.6012 \\
        \midrule
        C$^2$MIL ($\lambda_1:\lambda_2:\lambda_3=0.5:0.1:0.1$) & 0.6821&0.6520 &0.5979 \\
        C$^2$MIL ($\lambda_1:\lambda_2:\lambda_3=0.1:0.5:0.1$) & 0.6869& 0.6641& 0.5902\\
        C$^2$MIL ($\lambda_1:\lambda_2:\lambda_3=0.1:0.1:0.5$) & 0.6901& 0.6583& 0.5983\\
        \midrule
        C$^2$MIL ($\lambda_1:\lambda_2:\lambda_3=0.1:0.1:0.1$) & \textbf{0.7078} & \textbf{0.6904} & \textbf{0.6081} \\
        \bottomrule
      \end{tabular}
  }
  \caption{Ablation study (rows 1-3) and hyperparameter analysis (rows 4-6) in C$^2$MIL. The hyperparameters in the last row are the optimal ones chosen by C$^2$MIL. The hyperparameters are fixed at the optimal ones in the ablation study.}
  \label{tab:ada_vs}
\end{table}

\noindent \textbf{Optimization Strategies Analysis.}
Table \ref{tab:ada_vs} presents the ablation analysis of the optimization objectives. As shown in the table, removing any of the optimization objectives leads to a decline in the model's predictive performance. Notably, the removal of contrastive loss, which supervises the generation of causal subgraphs, has the most significant impact on performance, resulting in reductions of 2.99\%, 6.52\%, and 1.39\% across different datasets. This highlights the importance of the contrastive mechanism for causal subgraph learning.
In addition, we analyze the weighting of the optimization objectives. With weights $\lambda_1 = \lambda_2 = \lambda_3 = 0.1$, C$^2$MIL achieves the optimal performance while preventing excessive interference with survival primary optimization. 

\begin{figure}[htbp]
    \centering
    \includegraphics[width=0.5\textwidth]{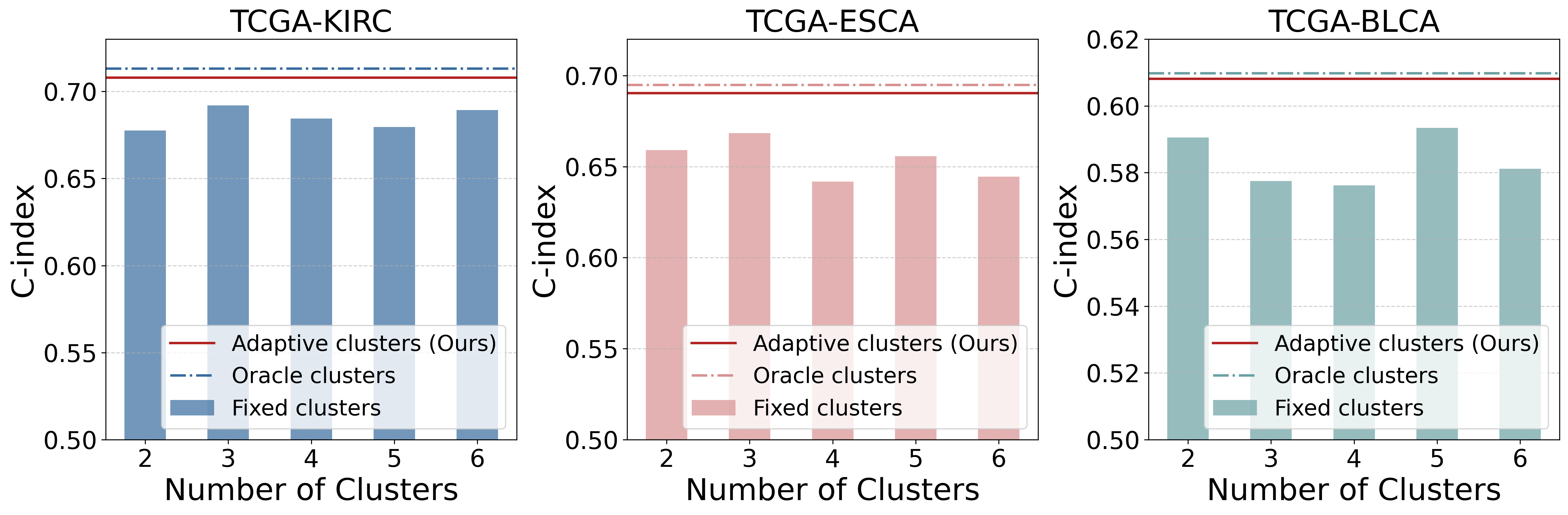}
    \caption{Predictive performance of adaptive (solid lines) vs. fixed (bars) and oracle (dashed lines) clustering. }
    \label{fig:clusterk}
\end{figure}

\noindent \textbf{Adaptive Cluster Number Analysis.}
\label{sec:clusterk}
In contrast to traditional approaches, C$^2$MIL adapts the number of clusters dynamically instead of employing a predetermined number of clusters, denoted as $K$. This experiment compares our adaptive clustering approach with those methods using a fixed number of clusters. Figure \ref{fig:clusterk} presents the five-fold cross-validation average results under different cluster settings (see Table \ref{tab:k_comparison} of Supplementary Material 
). The oracle clusters setting represents an ideal scenario where the best-performing cluster number $K$ is selected on test data for each fold, and the final result is averaged across five folds.

The empirical findings indicate that the utilization of an adaptive value for $K$ confers greater flexibility and results in enhanced predictive efficacy. Specifically, the performance of the model with adaptive $K$ either meets or exceeds that of models equipped with a fixed $K$, obviating the necessity for manual determination of the optimal cluster count. Moreover, the adaptive strategy closely aligns with the performance outcomes observed in the oracle clusters scenario. This inherent adaptability of the model enables it to estimate the number of clusters in response to the variances in the distribution of the training data, yielding exceptional predictive accuracy and enhanced computational efficiency.

\begin{figure}[htbp]
    \centering

    \includegraphics[scale=0.5]{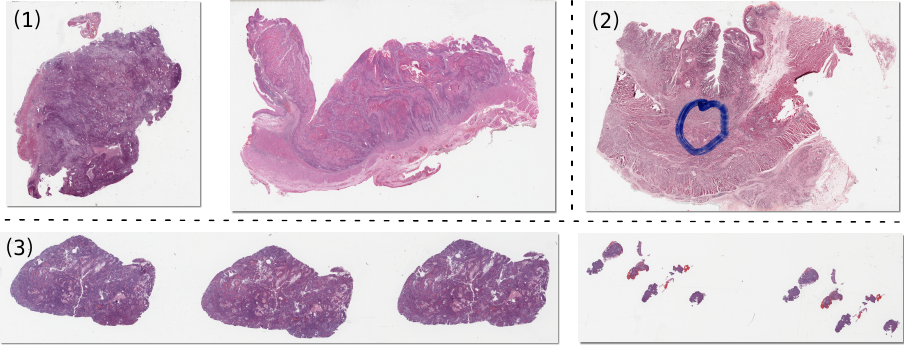}
    \caption{Visualization of thumbnail clusters in TCGA-ESCA test set, reflecting the overall irrelevant semantic bias.}

    \label{fig:cluster}
\end{figure}
\subsection{Interpretable Visualization}

\noindent \textbf{Adaptive Clustering Visualization.}
Figure \ref{fig:cluster} shows an example of the clustering results of thumbnails of one fold in the TCGA-ESCA. The C-index on the test set of this fold is improved from 0.7213 to 0.8214. The clustering results show that the model has adaptively learned three distinct categories, capturing differences in preparation methods. Cluster (1) tends to correspond to single-slide scenarios, Cluster (2) encompasses slides with annotations, and Cluster (3) is more likely to represent preparation methods involving multiple slices within a single file. Our method effectively hierarchizes the data and learns the preparation-specific trivial features.
\begin{figure}[htbp]
    \centering
    \includegraphics[scale=0.5]{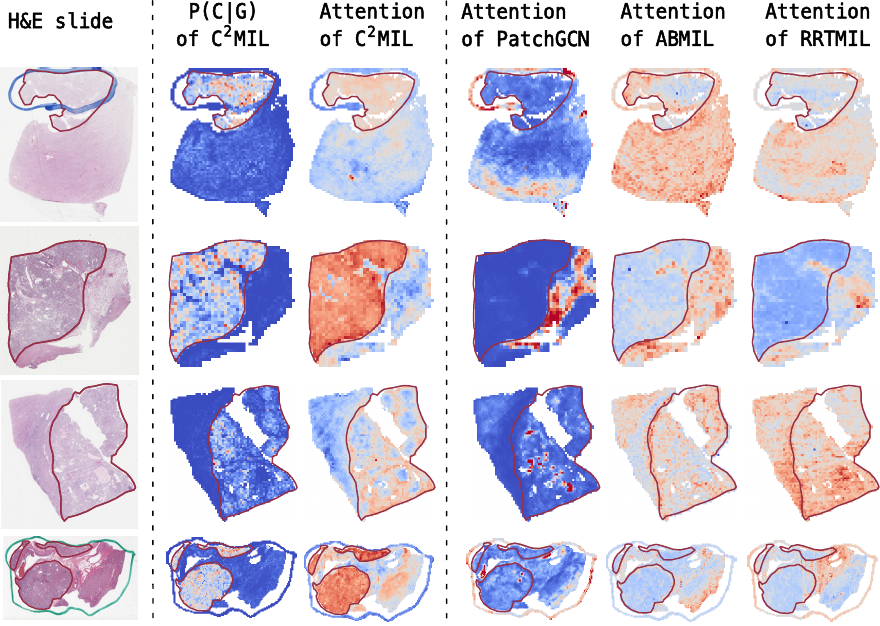}
    \caption{Attention heatmaps with cool-warm color bar. Region in red lines is ground truth.}
    \label{fig:attention}
\end{figure}

\noindent \textbf{Attention Heatmaps.}
Figure \ref{fig:attention} presents the heatmaps of estimation of $P(C\mid G)$  and the attention maps generated by C$^2$MIL of on TCGA-KIRC examples compared with ABMIL, PatchGCN, and RRTMIL. Unlike other models that highlight irrelevant or ambiguous regions, our model's high-attention areas (warm colors) align well with the pathology-based ground truth (red contours). By leveraging dual causalities, C$^2$MIL produces cleaner, clearer contours that exclude label-irrelevant regions and remain robust to preparation noise, demonstrating significantly improved interpretability and reliability compared to state-of-the-art methods.

\section{Conclusion}
\label{sec:conclusion}
We propose a dual-causal graph-based MIL framework, C$^2$MIL, in this paper to enhance accuracy, generalization, and interpretability by eliminating semantic confounders and identifying causal substructures. Experiments on public datasets achieve a state-of-the-art performance and 
an improved interpretability. 
C$^2$MIL shows strong potential for broader graph-based applications beyond survival analysis, including classification and detection. In future work, we plan to extend our approach to more general MIL settings to improve robustness and interpretability, especially for tasks where graph construction may not be feasible.

\textbf{Acknowledgements}. The work of Hong Zhang was partly supported by the National Natural Science Foundation of China (Grant No.7209121,12171451) and  Anhui Center for Applied Mathematics. This work of Liansheng Wang was supported by National Natural Science Foundation of China (Grant No. 62371409) and Fujian Provincial Natural Science Foundation of China (Grant No. 2023J01005).
{
    \small
    \bibliographystyle{ieeenat_fullname}
    \bibliography{main}
}

\clearpage
\setcounter{page}{1}
\maketitlesupplementary
\section{Method Supplementary}
\subsection{Graph Transformer Architecture Description}
\label{supp:GT}
Graph Transformer \cite{yun2019graph} consists of $L$ stacked identical layers, each containing multi-head graph attention mechanisms, positional encoding fusion, and position-enhanced feed-forward networks. The architecture is formally defined as follows:\\
\textbf{Input Representation.} Let graph $G=(V,E)$ contain $n$ nodes, where each node $i$ has feature vector $h_i \in \mathbb{R}^d$, with adjacency matrix $A \in \{0,1\}^{n\times n}$. The input feature matrix is $H^{(0)} = [h_1,\cdots,h_n]^T \in \mathbb{R}^{n \times d}$.\\
\textbf{Relative Posit Encoding.} The encoder structural relationship uses random walk probabilities:
\begin{equation}
    \mathbf{R}_{ij} = \text{Softmax}\left(\frac{\log(P_{ij})}{\sqrt{d}}\right),
\end{equation}
where $P \in \mathbb{R}^{n\times n}$ is the random walk transition probability matrix computed using k-step truncated values. \\
\textbf{Multi-head Graph Attention Mechanism.} For the h-th attention head in layer $l$:
\begin{equation}
    \begin{aligned}
        &\mathbf{Q}^{(h)} = \mathbf{H}^{(l)}\mathbf{W}_Q^{(h)},  \mathbf{K}^{(h)} = \mathbf{H}^{(l)}\mathbf{W}_K^{(h)},  \mathbf{V}^{(h)} = \mathbf{H}^{(l)}\mathbf{W}_V^{(h)},\\
        &\alpha_{ij}^{(h)} = \frac{\exp\left(\sigma\left(\frac{\mathbf{Q}_i^{(h)}(\mathbf{K}_j^{(h)})^\top}{\sqrt{d/H}} + \phi(A_{ij})\right)\right)}{\sum_{k \in \mathcal{N}_i} \exp\left(\sigma\left(\frac{\mathbf{Q}_i^{(h)}(\mathbf{K}_k^{(h)})^\top}{\sqrt{d/H}} + \phi(A_{ik})\right)\right)},
    \end{aligned}
\end{equation}
where $\phi: \mathbb{R} \rightarrow \mathbb{R}$ is an edge information mapping function, $\sigma$ denotes LeakyReLU activation, and $H$ is the number of attention heads.\\
\textbf{Structure-Aware Attention Aggrgation.} 
\begin{equation}
    \mathbf{Z}^{(h)} = \text{Softmax}(\boldsymbol{\alpha}^{(h)})\mathbf{V}^{(h)} + \mathbf{R} \circ (\boldsymbol{\alpha}^{(h)}\mathbf{V}^{(h)}),
\end{equation}
where $o$ denotes the Hadmard product. The multi-head output is concatenated:
\begin{equation}
    \hat{\mathbf{H}}^{(l)} = \|_{h=1}^H \mathbf{Z}^{(h)}\mathbf{W}_O^{(h)}.
\end{equation}
\textbf{Residual Connection \& Layer Normalization.} 
\begin{equation}
    \bar{\mathbf{H}}^{(l)} = \text{LayerNorm}\left(\mathbf{H}^{(l)} + \hat{\mathbf{H}}^{(l)}\right).
\end{equation}
\textbf{Postition-Enhanced Feed-Forward Network.}
\begin{equation}
    \small
    \mathbf{H}^{(l+1)} = \text{LayerNorm}\left(\bar{\mathbf{H}}^{(l)} + \mathbf{W}_2 \cdot \text{GELU}(\mathbf{W}_1\bar{\mathbf{H}}^{(l)} + \mathbf{b}_1) + \mathbf{b}_2\right).
\end{equation}
where $W_1 \in \mathbb{R}^{4d\times d}$ and $W_2 \in \mathbb{R}^{d \times 4d}$ are learnable parameters.
\\
\textbf{Output Layer}
Final node representations are obtained via K-hop neighborhood pooling:
\begin{equation}
    \mathbf{y}_i = \sum_{k=0}^K \gamma_k \cdot \text{MEAN}\left(\{\mathbf{H}_j^{(L)} | j \in \mathcal{N}_k(i)\}\right),
\end{equation}
where $\eta_k$ are learnable decay coefficients.
\subsection{Subgraph Sampling Pseudocodes}
\label{sup:ssp}
\begin{algorithm}[h]
\caption{Subgraph Sampling}
\label{alg:infer}
\begin{algorithmic}[1]
\Statex \textbf{Input:} Adjusted graph $G(\tilde{V},E,A)$; Linear $\text{MLP}(\cdot)$; Graph Transformer Model $GT(\cdot)$; $\text{subgraph}(\cdot,\cdot)$ function of graph containing the mask nodes; Activation function sigmoid $\sigma(\cdot)$.
\Statex \textbf{Output:} Causal graph $C$ and non-causal graph $S$.
\State $G'.V = \text{MLP}(G.V)$
\State $G'.E = G.E$
\State $G'.A = [G'.V_i; G'.V_j] \langle i,j\rangle \in G.E$
\State $P = \sigma(GT(G'))$
\If{training stage}
    \State sample = Bernoulli$(P)$
    \State mask = sample.detach() $+ P - P$.detach()
    \State $C = \text{subgraph}(G', \text{mask})$
    \State $S = \text{subgraph}(G', 1-\text{mask})$
\Else
    \State $C = \text{subgraph}(G', P)$
    \State $S = \text{subgraph}(G', 1-P)$
\EndIf
\end{algorithmic}
\end{algorithm}

\section{Experiments Supplementary}
\subsection{Implement Details}
\label{sup:Implement}
A pretrained UNI is used to extract features from both thumbnails and patches. The thumbnails are derived from WSIs at $40 \times$ magnification with a $30 \times$ downsampling. The patches are obtained by segmenting WSIs at $40 \times$ magnification into images of size $1024 \times 1024$ pixels. Before being fed into the feature extractor, both thumbnails and patches are resized to $224 \times 224$. Patches in a WSI is constructed as a graph by K nearest neighborhood (KNN) through the coordinates of patches. The proposed framework is implemented with PyTorch \cite{paszke2019pytorch} and PyTorch Geometric \cite{fey2019fast} and all the experiments are conducted on one NVIDIA A100 GPU with 40GB memory with batch size 16 and 100 epochs. The warm-up epoch is 2 on internal experiments and 10 on external experiments.


\subsection{Results of Adaptive Cluster Number (K) Analysis}
\label{supp:clusterk}
Specific value in Section \ref{sec:clusterk}

\begin{table}[h!]
  \centering
  \renewcommand{\arraystretch}{1.25}
  \setlength{\tabcolsep}{2pt}
  \resizebox{0.5\textwidth}{!}{
  \begin{tabular}{l|ccc}
    \toprule
    & TCGA-KIRC & TCGA-ESCA & TCGA-BLCA \\
    \hline
    $K=2$ & 0.6775 & 0.6591 & 0.5905 \\
    $K=3$ & 0.6920 & 0.6684 & 0.5775 \\
    $K=4$ & 0.6844 & 0.6418 & 0.5762 \\
    $K=5$ & 0.6795 & 0.6557 & 0.5934 \\
    $K=6$ & 0.6893 & 0.6445 & 0.5812 \\
    \hline
    Adaptive clusters (Ours) & \textbf{0.7078} & \textbf{0.6904} & \textbf{0.6081} \\
    \hline 
    Oracle clusters & 0.7131 & 0.6949 & 0.6098 \\
    \bottomrule
  \end{tabular}
  }
  \caption{Predictive performance analysis of the adaptive optimal clustering number method compared with fixed number $K$ of clusters.}
  \label{tab:k_comparison}
\end{table}


\end{document}